%% file: PaperForReview.tex
\documentclass[10pt,twocolumn,letterpaper]{article}

\usepackage{cvpr}              

\usepackage{graphicx}
\usepackage{amsmath}
\usepackage{amssymb}
\usepackage{booktabs}

\usepackage[pagebackref,breaklinks,colorlinks]{hyperref}
\usepackage{multirow}
\usepackage{makecell}
\usepackage{enumerate}

\usepackage[capitalize]{cleveref}
\crefname{section}{Sec.}{Secs.}
\Crefname{section}{Section}{Sections}
\Crefname{table}{Table}{Tables}
\crefname{table}{Tab.}{Tabs.}

\def\cvprPaperID{7326} 
\def\confName{CVPR}
\def\confYear{2023}

\input{symbols}

\begin{document}

\title{Semi-Weakly Supervised Object Kinematic Motion Prediction}


\author{Gengxin Liu\textsuperscript{\rm1} ~~~
Qian Sun\textsuperscript{\rm1} ~~~ 
Haibin Huang\textsuperscript{\rm2} ~~~ 
Chongyang Ma\textsuperscript{\rm2} ~~~ 
Yulan Guo\textsuperscript{\rm3} ~~~ 
\\
Li Yi\textsuperscript{\rm4}~~~
Hui Huang\textsuperscript{\rm1} ~~~
Ruizhen Hu\textsuperscript{\rm1}\thanks{Ruizhen Hu is the corresponding author.}
\\
\textsuperscript{\rm1}Shenzhen University 
\textsuperscript{\rm2}Kuaishou Technology 
\textsuperscript{\rm3}Sun Yat-sen University
\textsuperscript{\rm4}Tsinghua University
\\
{\tt\small \{gengxin.v.liu,qiansun1006,jackiehuanghaibin,chongyangm,yulan.vision,}
\\
{\tt\small ericyi0124,hhzhiyan,ruizhen.hu\}@gmail.com }
}

\maketitle
\newcommand{\cmt}[1]{\textcolor{red}{{[Comments: #1]}}}
\newcommand{\todo}[1]{\textcolor{blue}{{[todo: #1]}}}
\newcommand{\rh}[1]{\textcolor{blue}{{#1}}}
\newcommand{\gx}[1]{\textcolor{red}{{#1}}}

\input{abs}

\input{intro}
\input{related_work}

\input{method}

\input{experiments}

\input{conclusion}

\section{Acknowledgement}
We thank the reviewers for their valuable comments. 
This work was supported in parts by NSFC (U2001206, U21B2023, U20A20185, 61972435), GD Natural Science Foundation (2021B1515020085, 2022B1515020103), GD Talent Program (2019JC05X328), Shenzhen Science and Technology Program (RCYX20210609103121030), DEGP Innovation Team (2022KCXTD025), and Guangdong Laboratory of Artificial Intelligence and Digital Economy (SZ).





{\small
\bibliographystyle{ieee_fullname}
\bibliography{egbib}
}


\end{document}

%% file: symbols.tex
\newcommand{\loss}{\mathcal{L}}
\newcommand{\typeloss}{\loss_\mathrm{cls}}
\newcommand{\dirloss}{\loss_\mathrm{dir}}
\newcommand{\posloss}{\loss_\mathrm{pos}}

%% file: abs.tex
\begin{abstract}
Given a 3D object, kinematic motion prediction aims to identify the mobile parts as well as the corresponding motion parameters. Due to the large variations in both topological structure and geometric details of 3D objects, this remains a challenging task and the lack of large scale labeled data also constrain the performance of deep learning based approaches. In this paper, we tackle the task of object kinematic motion prediction problem in a semi-weakly supervised manner. Our key observations are two-fold. First, although 3D dataset with fully annotated motion labels is limited, there are existing datasets and methods for object part semantic segmentation at large scale. Second, semantic part segmentation and mobile part segmentation is not always consistent but it is possible to detect the mobile parts from the underlying 3D structure. Towards this end, we propose a graph neural network to learn the map between hierarchical part-level segmentation and mobile parts parameters, which are further refined based on geometric alignment. This network can be first trained on PartNet-Mobility dataset with fully labeled mobility information and then applied on PartNet dataset with fine-grained and hierarchical part-level segmentation. The network predictions yield a large scale of 3D objects with pseudo labeled mobility information and can further be used for weakly-supervised learning with pre-existing segmentation. Our experiments show there are significant performance boosts with the augmented data for previous method designed for kinematic motion prediction on 3D partial scans.
\end{abstract}

%% file: intro.tex
\section{Introduction}
\label{sec:intro}
In this work, we study the problem of 3D object kinematic motion prediction. As a key aspect of functional analysis of 3D shape, the understanding of part mobilities plays an important role in 
applications, such as robot-environment interaction, object functionality prediction, as well as 3D object reconstruction and animation.

There have been a few methods on discovering part mobility from various inputs, including 3D objects and sequence of RGBD scans, via both supervised or unsupervised learning~\cite{hu2017learning,wang2019shape2motion,yan2020rpm,kawana2021unsupervised,xu2022unsupervised}. 
Despite the rapid development in this filed, it remains challenging to boost the generalization performance of these methods: there is significant performance degeneration when the queried objects present different topology, structure, or geometric details that are out of the space of training examples.

\input{figures/teaser.tex}

However, it is with great difficulty to collect fully annotated dataset for kinematic motion prediction, which requires not only segmentation of object w.r.t motion but also the associated parameters.
Furthermore, it is infeasible to collect such dataset at large scale, since there are only a few dataset with limited size available for motion prediction task such as PartNet-Mobility~\cite{Xiang_2020_SAPIEN} and Shape2Motion~\cite{wang2019shape2motion}. 
Meanwhile, as a long lasting task in both graphics and vision, there have been massive methods proposed with large scale dataset of segmented 3D objects. Among them, PartNet dataset~\cite{mo2019partnet} consists of more than 25k 3D models of 24 object categories with the corresponding fine-grained and hierarchical 3D part information.
Given such a dataset that covers sufficient variations in 3D topological structure and geometric details, we wonder if we can bridge the best of two worlds: transfer the fully labeled information from PartNet-Mobility into PartNet, and leverage the augmented data for motion prediction learning in a semi-weakly supervised manner, as shown in Figure~\ref{fig:teaser}. 

Such transfer is non-trivial and the main challenge lies in the fact that existing segmentation of a 3D model does not necessarily comply with its mobilities.
For example, for most car models, the door is not a separate component.
This fact significantly increases the difficulty of transfer: it is not just adding mobility parameters to the given segmentation but also to find the appropriate mobile parts.

Since both PartNet-Mobility and PartNet provide fine-grained and hierarchical part-level segmentation, we unitize this information and convert the mobile part detection task into a part selection task from the given hierarchy.
Specifically, we develop a two-stage pipeline in a coarse-to-fine manner.
In the first stage, we construct a graph that encodes both the parent-child hierarchy and adjacent part pairs. We then train a graph neural network to predict the motion part at the edge level, including its motion type and corresponding parameters. It is trained with PartNet-Mobility data and then applied on PartNet to get the initial prediction.
In the second stage, we introduce a refinement process to filter out prediction with low feasibility score and consistency score, refine the motion axis using a heuristic strategy, and generate the final pseudo labels as motion prediction results.

To evaluate the effectiveness of our generated pseudo prediction, we further apply it on the task of motion prediction on partial scans. We adapt the ANCSH method proposed in \cite{li2020category} and enhance its training with the augmented data. Our experiments show that the prediction accuracy can be improved by a large margin and our approach outperforms state-of-the-art methods under such semi-weakly supervised learning pipeline. 

To summarize, our contributions are as follow:
\begin{itemize}
\item We are the first to tackle the problem of object kinematic motion prediction using a semi-weakly supervised learning framework. Specially, we successfully transfer the fully labeled motion information from a dataset of limited examples to a large set of unlabeled 3D shapes by leveraging the pre-existing segmentation hierarchy information. The augmented data can be used for weakly supervised learning at large scale.      

\item  We propose a robust two-stage pipeline for motion information transfer: A 
graph neural network is first adapted to train and predict the map between semantic segmentation hierarchy and part motion properties. Then a heuristic strategy is presented to refine and output the final predictions. 

\item We further evaluate our pipeline on the task of 3D partial scans motion prediction. Our experiments show that existing approaches trained with the data augmented by our method can be improved significantly in terms of prediction accuracy.  
\end{itemize}

%% file: figures/teaser.tex
\begin{figure}
    \centering

    \includegraphics[width=\linewidth]{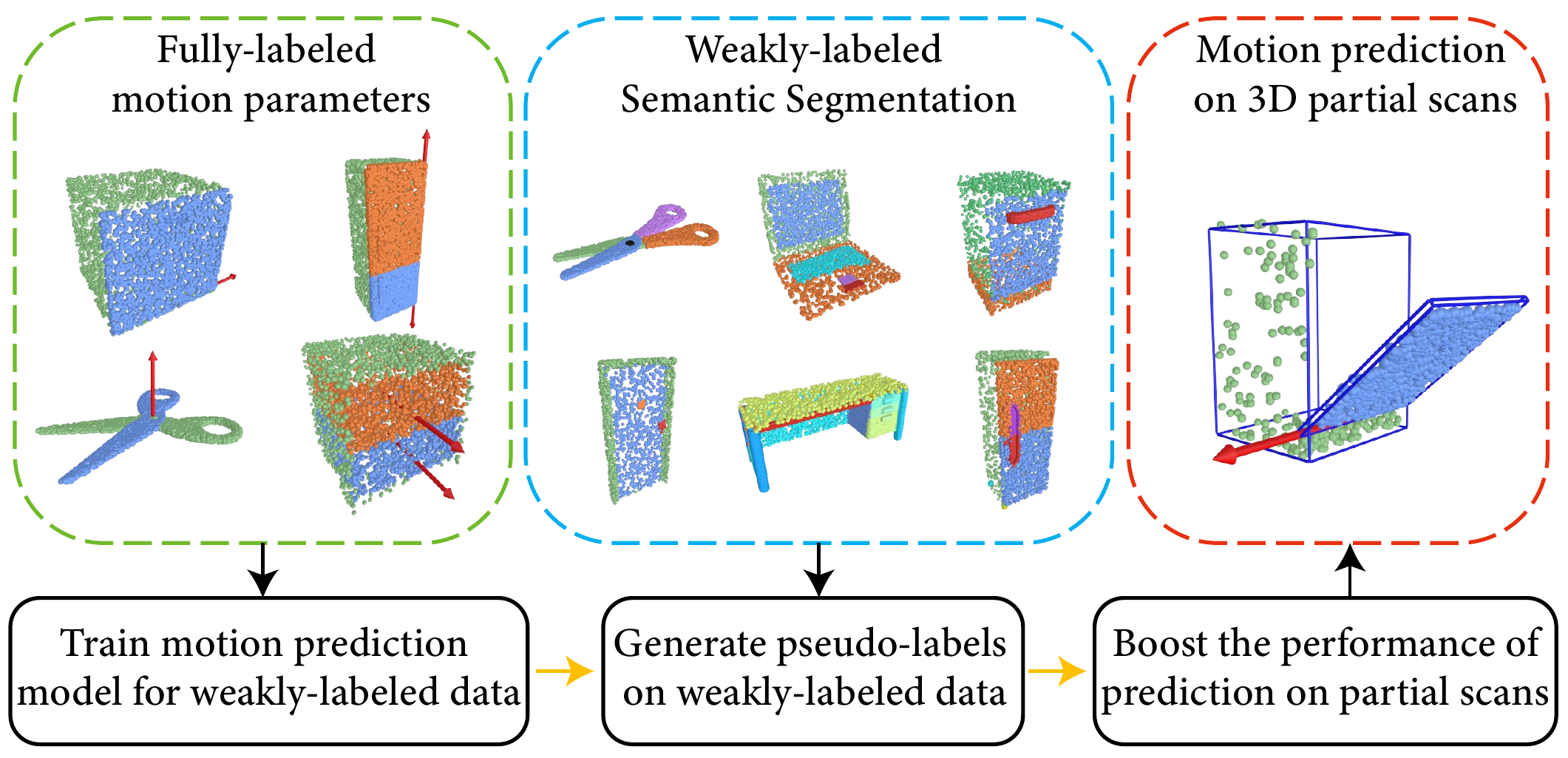}
    
\caption{Our semi-weakly supervised method for object kinematic motion prediction, which utilize the weakly labeled dataset with part semantic segmentation to augment the training data for object kinematic motion prediction on partial scans.}
\label{fig:teaser}
\end{figure}

%% file: related_work.tex
\section{Related Work}
\label{sec:related}

\paragraph{Object kinematic motion prediction.}
Early work with traditional method for kinematic motion prediction includes slippage analysis for deformable mesh models \cite{xu2009joint}, illustrating the motions of mechanical assemblies \cite{mitra2010illustrating}, and inferring kinematic chains based on motion trajectories \cite{yan2006automatic}. 
Recently, machine learning based approaches have been applied to point clouds to predict their motions.
Hu et al.\cite{hu2017learning} introduces a learning method for the mobility of parts in 3D
objects, which assumes consistently-segmented manufactured object meshes are given. Wang et al.\cite{wang2019shape2motion} takes a single 3D point cloud as input, and jointly computes a mobility-oriented segmentation and the associated motion attributes. Yan et al.\cite{yan2020rpm} simultaneously infers movable parts and hallucinates their motions from a single un-segmented 3D point cloud.
On the other hand, Kawana et al \cite{kawana2021unsupervised} propose an unsupervised pose-aware part decomposition method to explicitly target man-made articulated objects with mechanical joints. While their method requires many pose variations of each shape for training. Xu et al. \cite{xu2022unsupervised} introduces the concept of category closure, and their method requires only one observation of each unique shape for training. Although these methods achieve promising results for motion prediction under their setting, there is significant performance degeneration for 3D objects with complicate structure due to the limited data used for training.  Hence, we propose a semi-weakly supervised learning approach that extend the limited fully labeled data into a large scale 3D object repository and improve the performance. 


\input{figures/overview.tex}

\paragraph{Semi-weakly supervised learning.}
As a popular learning framework in computer vision,  semi-supervised learning aim to learn from both labeled and unlabeled data  and gain performance improvements over learning on labeled data only \cite{lee2013pseudo, zou2018unsupervised, iscen2019label, arazo2020pseudo}. Partially, these kind of methods first train a supervised model with labeled data and then predict pseudo-labels in unlabeled data, the combined data is then used to train the final model. 
In addition, weakly supervised learning leverage weak supervision on unlabeled data to boost the performance of the supervised method. For example, the weakly supervised object detection in computer vision often rely on image-class labels \cite{bilen2016weakly, diba2017weakly, tang2017multiple}, and weakly supervised image segmentation leverages the image tags \cite{chang2020weakly}, bounding boxes \cite{song2019box}, labeled points and scribbles \cite{tang2018regularized} to improve the segmentation performance. 
Our method work in both semi- and weakly-supervised manner.  We first use the fully labeled motion in PartNet-Mobility dataset to train a graph neural network where we leverage the fine-grained and hierarchical part-level segmentation as weak label. We applied the trained graph neural network to predict pseudo-label motion in PartNet. Finally, we use the pseudo-labeled motion to boost the performance for motion prediction on 3D partial scans. We demonstrate in experiment that pure semi-supervised learning have limited performance improvement and even may hurt the model.

%% file: figures/overview.tex
\begin{figure*}
    \centering
    \includegraphics[width=\textwidth]{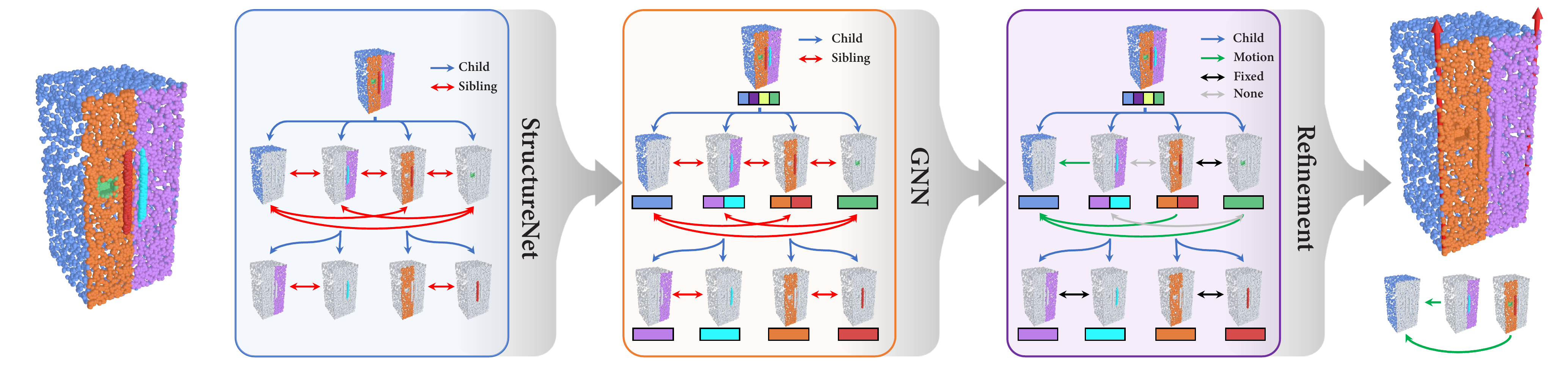}
\caption{Overview of our method. Given an input shape with fine-grained hierarchical part-level segmentation, we first use the StructureNet~\cite{mo2019structurenet} to  encode and propagate features for all the part nodes in the hierarchy. Then we use a graph neural network \cite{scarselli2008graph} to predict the motion parameters for edges connecting sibling part nodes in the hierarchy. After predicting the results on all the edges, we prune the hierarchy and extract the final mobility tree, and further leverage geometric priors to refine the motion parameters.}
\label{fig:overview}
\end{figure*}

%% file: method.tex
\section{Our Method}
\label{sec:method}


Our goal is to use the PartNet-Mobility dataset~\cite{Xiang_2020_SAPIEN} with fully labeled mobility information to train a network to predict the motion parameters for shapes in the PartNet dataset~\cite{mo2019partnet} with only fine-grained and hierarchical part-level segmentation, so that they can be also used to generate pseudo ground-truth data for motion prediction on partial scans. The overview of our method is shown in Figure~\ref{fig:overview}.

\subsection{Notations}

For each shape $S$, its segmented parts can be denoted as $P=\{P_i\}$ and the hierarchy can be denoted as $H \subset P^2$ consisting of directed edges from parent parts to all of their children. 
As our goal is to determine the relative motion between parts, and such motion can only exist for adjacent part pairs, so other than the above parent-child edges, we further construct a set of edges between adjacent parts decoded as $A \subset P^2$ and our goal becomes to classify the motion type of edges in $A$ as well as predict detailed motion parameters between part pairs with relative motion.

For the motion parameters, we use the fine-grained motion types following~\cite{hu2017learning}, where the classification is based on the joint type (prismatic, revolute, or both, denoted as \emph{P}, \emph{R}, and \emph{PR}), the general direction of the axes (horizontal or vertical, denoted as \emph{H} and \emph{V}), as well as the location of the pivot point for revolute joint (close to the center of the moving object or to one of the sides, denoted as \emph{C} or \emph{S}).
For part pairs that do not have relative motion, they could be either in fixed connection or isolated, so we add another two types of relation types, which results in 10 different edge types, including \emph{P\_H, P\_V, R\_H\_C, R\_H\_S, R\_V\_C, R\_V\_S, PR\_H, PR\_V, Fixed,} and \emph{None}. 
The motion parameter for each part pair is demoted as $m=(t,d,p)$ where $t$ is the motion type, $d$ and $p$ are the direction and position of the motion axis, respectively.
Note that for relation types \emph{Fixed} and \emph{None}, we ignore both parameters $d$ and $p$, while for relation types \emph{P\_H} and \emph{P\_V}, we ignore the parameter $p$.

\subsection{Network Optimization}

Given an input shape $S$ consisting of parts $P=\{P_i\}$ in a hierarchy $H$ with the relationships between adjacent parts denoted as $A = \{a_k\}$ , our goal is to predict the motion parameters $m_k$ for each edge $a_k \in A$.

We first get two types of part-level representations.
For each leaf part $P_i$, we sample 1024 points for detailed geometry representation $G_i$ used as input for node-level feature extraction.
For each part $P_i$ including the intermediate node in the Hierarchy, we further compute its symmetry-aligned oriented bounding box (OBB) $B_i$ as in~\cite{fish2014meta}.
Then we pass the point sampling of all the leaf nodes $\{G_i\}$  to the StructureNet~\cite{mo2019structurenet} to propagate and encode features for all the part nodes in the hierarchy. 
We then encode the relative position between two OBBs for each adjacent sibling part pair as the initial edge-level feature.
A graph neural network (GNN)~\cite{scarselli2008graph} is then used to predict the motion parameters for each sibling $a_k \in A$.

The loss function of the prediction for each part pair is defined as 
\begin{equation}
    \loss = \typeloss(t, \hat{t}) + \dirloss(d, \hat{d}) + \posloss(p, \hat{p})
\end{equation}
where $\hat{t}$, $\hat{d}$, and $\hat{p}$ are the ground-truth edge types, joint direction, and joint position, respectively.
For edge types loss $\typeloss$, we use the softmax cross-entropy loss:
\begin{equation}
    \typeloss(t, \hat{t}) =  -\sum_{i=1}^{10}{\hat{t}_{i} \log{(t_i)} }
\end{equation}
where $\hat{t}_i$ is the ground-truth label and $t_i$ is the softmax probability for the $i$th class.
We use the mean-squared error for the joint direction loss $\dirloss$:
\begin{equation}
    \dirloss(d, \hat{d}) =  \lVert d- \hat{d} \rVert .
\end{equation}
Finally, the loss term for joint positions is defined as
\begin{equation}
    \posloss(p, \hat{p}) =  \frac{\lVert (p- \hat{p}) \times \hat{d} \rVert}{\lVert \hat{d} \rVert },
\end{equation}
where $\times$ denotes the cross product.

\subsection{Refinement of Prediction Results}
Once we have the prediction results on all the edges, we prune the hierarchy to leave only the nodes that participant the motion. For motion-related parts connected with fixed edge but point to the same node, i.e., take the same part as the reference part, we merge them into one node as movable part and take the motion parameter of the larger part.

With the final mobility tree, we can directly use the motion parameters predicted for the corresponding motion edge. However, we found that the prediction is inaccurate due to the limited training data. Based on the key observation that most of the motion axes is parallel to or lie in the edges of the OBBs of the parts, we proposed to select the final motion axes from the candidate axes generated from the OBBs based on the feasibility score and consistency score. 

Specifically, we first select a set of candidate motion axes $\{m_i\}$ from the movable part's OBB according to the predicted motion type $t$, based on whether the axis is horizontal or vertical and whether the axis is on the side or in the central region of the part. 
For axis on the central region of the part, other than taking the OBB centroid as the pivot point, we further compute the interaction region~\cite{hu2015interaction} between the movable part and the reference part and take its centroid as another pivot point to form an additional candidate axis.

\input{figures/tab_refinement.tex}

To measure the feasibility, we train an axis selection network to predict a confidence score $s_i^f$ for each candidate axis $m_i$.  
The axis selection network is inspired from the Siamese network \cite{bromley1993signature}.
The input to the network includes the original point clouds and the transformed point clouds after applying the candidate motion $m_i$, associated with point-wise one-hot label indicating whether the point belongs to the movable or reference part, which are then passed to a DGCNN encoder~\cite{wang2019dynamic} to extract the global feature of each point cloud.
Those two global features are then concatenated and fed into fully connected layers to output a feasibility score for the relative motion.
To train the axis selection network, we use the ground-truth motion parameter 
to generate positive transformed point clouds, with feasibility score as 1, and sample a random edge of the movable part's OBB that is different from the GT axis for negative examples generation, with feasibility score as 0. 
The network is then trained using a binary cross entropy loss.

Other than the feasibility, we also compute the consistency between the candidate motion $m_i$ and predicted motion from our network $m = \{t, d, p\}$ based on the direction similarity $s_i^d$ and position similarity $s_i^p$ as follows: 
\begin{equation}
    s_i^d = \frac{ d \cdot d_i }{ \lVert d \rVert \lVert d_i^c \rVert } 
\end{equation}
\begin{equation}
    s_i^p = 1 - \frac{\lVert (p - p_i) \times p \rVert}{\lVert p_i \rVert }.
\end{equation}

Then we compute the final score $s_i$ of each candidate motion parameter $m_i$, and select the best candidate motion with the highest score:
\begin{equation}
    s_i = w_f s_i^f + w_d s_i^d + w_p s_i^p
\label{eq:motionscre}
\end{equation}
where $w_f=0.5$, $w_d=0.3$, and $w_p=0.2$ are the corresponding weights.

\subsection{Motion Prediction on Partial Scans}
\label{sec:ancsh}
For motion prediction on partial scans, we adopt the method ANCSH proposed in~\cite{li2020category}, where a canonical representation is introduced to addresses the task of category-level pose estimation for articulated objects from a single depth image. 
The input to ANCSH network is a depth point cloud, and the output of the network are rigid part segmentation, dense coordinate predictions in each Normalized Part Coordinate Spaces (NPCS), transformations from each NPCS to Normalized Articulated Object Coordinate Space (NAOCS), as well as joint parameters in NAOCS, correspondingly. 
Then ANCSH uses RANSAC \cite{fischler1981random} based framework to estimate the 6D pose and size of each rigid part. 
Finally, ANCSH computes the joint parameters and joint states in the camera-space based on the part pose and size. 
To generate the training and testing data for the ANCSH network, PyBullet \cite{michalik2020pybullet} is used to render depth image with different articulations from different viewpoints. 

To boost the performance of ANCSH, we first use the score computed in Eq.~\ref{eq:motionscre} to automatically filter out those samples with poor prediction quality and leave the ones with score larger than a given threshold $\tau=0.6$, and then follow the same rendering pipeline to generate augmented data for training.

%% file: figures/tab_refinement.tex
\begin{table*}[t!]
\addtolength{\tabcolsep}{-2pt}
\centering
\begin{tabular}{c|cccc|cccc}\toprule
                   & \multicolumn{4}{c|}{Angle Error°↓ }                     & \multicolumn{4}{c}{Distance Error↓ }                        \\ \midrule 
Method             & OBB\_fine      & GNN\_fine & OBB\_coarse    & GNN\_coarse   & OBB\_fine      & GNN\_fine      & OBB\_coarse    & GNN\_coarse \\ \midrule 

refrigerator       & \textbf{0.24}  & 3.22      & \textbf{0.24}  & 7.30          & 0.139          & \textbf{0.125} & 0.130          & 0.239       \\
display            & \textbf{12.05} & 14.32     & 69.72          & 36.45         & 0.172          & \textbf{0.081} & 0.214          & 0.089       \\
laptop             & \textbf{0.48}  & 1.72      & \textbf{0.48}  & 3.31          & 0.044          & 0.070          & \textbf{0.036} & 0.079       \\
Knife              & \textbf{8.63}  & 21.72     & \textbf{8.63}  & 30.79         & 0.573 & \textbf{0.493}          & 0.587          & 0.598       \\
clock              & \textbf{3.65}  & 3.95      & 30.39          & 4.95          & 0.273          & \textbf{0.028} & 0.133          & 0.036       \\
bottle             & \textbf{1.26}  & 4.69      & 30.50          & 5.50          & \textbf{0.012} & 0.094          & 0.043          & 0.216       \\
scissors           & \textbf{0.79}  & 2.06      & \textbf{0.79}  & 4.84          & \textbf{0.078} & 0.099          & 0.085          & 0.098       \\
table              & 0.84           & 0.62      & 7.35           & \textbf{0.32} & -              & -              & -              & -           \\
dishwasher         & \textbf{0.59}  & 1.76      & 30.06          & 3.61          & 0.260          & 0.197          & \textbf{0.137} & 0.146       \\
storage\_furniture & \textbf{10.89} & 11.77     & 13.80          & 13.34         & \textbf{0.174} & 0.278          & 0.266          & 0.213       \\
door\_set          & \textbf{8.60}  & 12.19     & 36.34          & 12.03         & 0.144          & 0.393          & \textbf{0.087} & 0.423       \\
Pliers             & \textbf{4.06}  & 4.80      & \textbf{4.06}  & 11.19         & 0.005          & \textbf{0.002} & 0.066          & 0.044       \\
Pen                & \textbf{15.00} & 22.00     & 27.77          & 24.55         & -              & -              & -              & -           \\
Stapler            & \textbf{1.04}  & 6.76      & 1.05           & 7.43          & \textbf{0.131} & 0.154          & 0.144          & 0.152       \\
Oven               & \textbf{8.28}  & 11.47     & 48.84          & 20.00         & \textbf{0.024} & 0.340          & 0.033          & 0.419       \\
Luggage            & \textbf{0.08}  & 17.37     & 45.07          & 24.40         & 0.071          & 0.170          & \textbf{0.052} & 0.416       \\
Window             & \textbf{0.12}  & 8.17      & \textbf{0.12}  & 3.61          & -              & -              & -              & -           \\
Cart               & \textbf{7.64}  & 6.63      & 7.65  & 8.04          & 0.070 & 0.206          & \textbf{0.041}          & 0.403       \\
USB                & 11.24          & 20.61     & \textbf{11.22} & 20.17         & \textbf{0.012} & 1.252          & 0.038          & 0.625       \\
FoldingChair       & \textbf{0.08}  & 2.70      & \textbf{0.08}  & 2.71          & \textbf{0.061} & 0.111          & 0.076          & 0.200       \\ \midrule 
mean               & \textbf{4.78}  & 8.93      & 18.71          & 12.23         & 0.112 & 0.205          & \textbf{0.108}          & 0.220      \\ \bottomrule 
\end{tabular}
\caption{Motion prediction results on pre-segmented shapes. The design of using fine motion types and OBB prior to refine motion parameters, i.e., OBB\_fine used in our method, leads to the best performance for most categories.}
\label{tab:result_seg}
\end{table*}

%% file: experiments.tex
\input{figures/fig_fine_direction.tex}
\input{figures/fig_fine_position.tex}

\section{Experimental Results}
\label{sec:exp}

In this section, after explaining the experiment setup in Section~\ref{sec:exp_metrics}, we will first justify our method design for motion prediction on pre-segmented shapes in Section~\ref{sec:exp_gnn}, and then show our semi-weakly supervised method can improve the motion prediction results on partial scans in Section~\ref{sec:exp_articulated}.



\subsection{Experimental metrics}
\label{sec:exp_metrics}
We use the following metrics as in \cite{li2020category} to evaluate our method. 
For joint parameters, we evaluate the orientation error of the motion axis in degrees for revolute joint and prismatic joint, and using the minimum line-to-line distance to evaluate the position error for revolute joint. 
For part prediction, we use the rotation error, translation error, and 3D intersection over union (IoU) as part-based metrics. 
For joint states, we evaluate joint angle error in degrees for each revolute joint, and evaluate the error of relative translation amounts for each prismatic joint. 


\subsection{Prediction Results on Pre-segmented Shapes}
\label{sec:exp_gnn}

To evaluate the motion prediction accuracy on the shapes from PartNet dataset that with fine-grained and hierarchical part-level segmentation, we test our method on a subset of PartNet-Mobility dataset with GT motion parameters. We found that motion type directly output by the GNN network is quite high and the main error comes with the axis parameters, and our key design is to use finer motion types and refine the axis using OBB candidates, so we mainly compare the performances of settings whether those two improvements are adopted, which results in four settings: \emph{GNN\_coarse, OBB\_coarse, GNN\_fine, OBB\_fine}. We measure angle error and distance error between the predicted axis and the GT axis, and report the results on 20 categories in Table~\ref{tab:result_seg}. Note that for prismatic joint, there is no pivot point and only angle error is computed.

We can see that when comparing the results using different sets of motion types, the angle error is always smaller with fine motion type, no matter the final prediction is directly from GNN or refined using OBB. This is because the fine characterization of the motion type provide more direction information of the axis, which provides more constraints on the predictions and leads to better result; see Figure~\ref{fig:result_fine_direction} for some visual comparisons.  

Moreover,  when comparing the results before and after refining using OBB, we  see that the distance error is consistently decreased, which indicates that using OBB priors can correct the inaccurate prediction directly output by the network; Figure~\ref{fig:result_fine_position} shows some visual comparisons.  
We  see that the motion position predicted by \emph{OBB\_fine} are more accurate than \emph{GNN\_fine} since the OBB priors and the interaction region can better locate the position of the joint. For example, the axis location of USB and folding chair in the first and second row are refined with the introduction of interaction region, while the axis location of bottle, cart and door in the next three rows
are refined by the OBB priors, which are more accurate than those directly predicted by \emph{GNN\_fine}. 
As a result, our strategy of using fine motion types and OBB priors, i.e., \emph{OBB\_fine},  gets the best result.

\input{figures/tab_semiweakly.tex}

\subsection{Prediction Results on Partial Scans}
\label{sec:exp_articulated}

When the input are partial scans, we show that our semi-weakly supervised method can highly boost the articulated pose estimation performance of 
\cite{li2020category}, denoted as \emph{ANCSH}, and our method is denoted as \emph{ANCSH (semi-weakly)}.

\textbf{Semi-weakly supervised vs. semi-supervised.}
To show that the weakly-labeled semantic segmentation has played an important role, we also compare the results obtained by the semi-supervised setting, denoted as \emph{ANCSH (semi)}. This semi-supervised baseline first trains the ANCSH model with the PartNet-Mobility dataset and then predicts the pseudo label on unsegmented PartNet dataset. The predicted pseudo label with high confidence is then selected to augment the training set to obtain the final results. 


\input{figures/fig_result_partial.tex}

\input{figures/tab_supervision.tex}

The quantitative comparisons are shown in Table~\ref{tab:result_semiweakly}. We see that when using our augmented data with weakly-labeled semantic segmentation, there are significant performance boosts on each category, while the improvement of ANCSH (semi) is relatively weak and sometimes even hurt the model. 
Some visual comparisons of results obtained in different settings are shown in Figure~\ref{fig:result_partial}.
We can see that \emph{ANCSH (semi-weakly)} predicts more accurate motion direction and position, such as the dishwasher and table in the second and the third row. Moreover, \emph{ANCSH (semi-weakly)} also outputs more clear part segmentation, such as the refrigerator and scissors in the first and the fifth row. Finally, amodal bounding boxes obtained by \emph{ANCSH (semi-weakly)} are also more accurate, such as the table and the door shown in the third and the fourth rows.

\textbf{Impact of the ratio parameter.}
We also investigate the effect of different ratios of training data on two categories. Specifically, we first use different ratios of training data on PartNet-Mobility dataset to train our model proposed in Section~\ref{sec:method}, and then use the model to predict pseudo ground truth in PartNet dataset. 
The same set of training data of PartNet-Mobility dataset together with the predicted augmented PartNet data is used to train ANCSH. 
Results are shown in Table~\ref{tab:result_supervision}. 
We can see that when the training data is 80\%, \emph{ANCSH (semi-weakly)} can still outperform \emph{ANCSH} that trained using 100\% training data, and \emph{ANCSH (semi-weakly)} is also comparable when the training data is down to 50\%. This demonstrates that the augmented data predicted by our method can help to boost the performance when the training data is limited. 

\textbf{ Robustness to training data.}
As our key contributions is to utilize the hierarchical part segmentation information for motion prediction, other than using the GT part segmentation, we further conduct our experiment on new shapes generated by the most recent work DSG-Net~\cite{yang2022dsg_rebuttal}. 
Note that all those shape are generated with associated hierarchical part segmentation. 
Since StorageFurniture is the only category they generated that contains movable parts, we show the results on this category in Table \ref{tab:result_dsg1}, where \emph{ANCSH} refers to the original fully-supervised setting and \emph{ANCSH (DSG)} refers to for our semi-weakly supervised setting using the inferred pseudo labels on shapes generated by DSG-Net. 
We can see that the augmented data from DSG-Net also boost the performance, which indicates the generality of our method and robustness to the training data without human-annotated part segmentation.
\input{figures/tab_dsg1.tex}


%% file: figures/fig_fine_direction.tex
\begin{figure}[!t]
    \centering
    \includegraphics[width=0.94\linewidth]{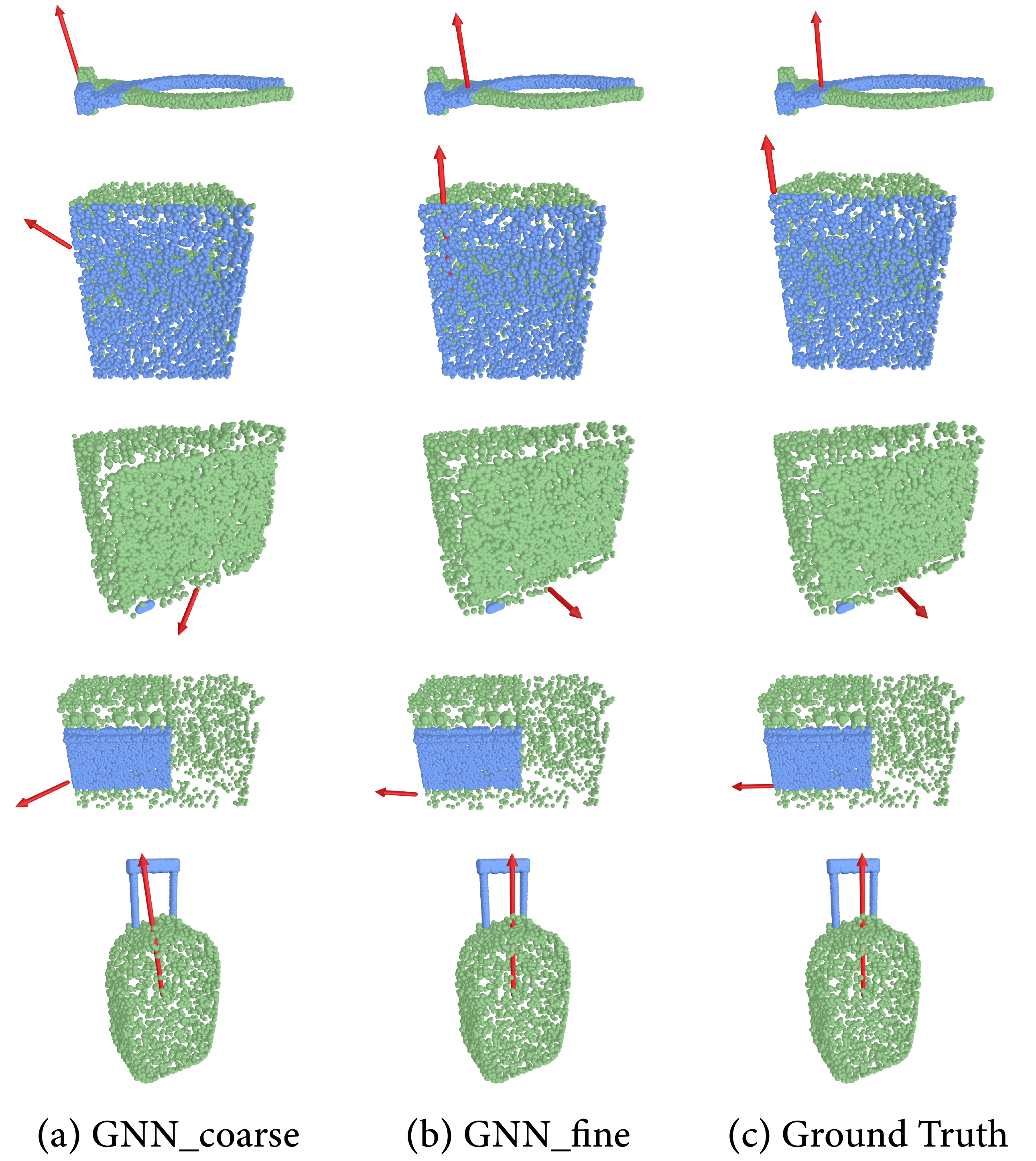}

\caption{Visual comparison of results obtained by  \emph{GNN\_coarse} and \emph{GNN\_fine} on pre-segmented point clouds. 
The focused movable part is colored in blue while all other parts are colored in green.
\emph{GNN\_fine} outputs more accurate axis direction owing to the fine characterization of the motion type.}
\label{fig:result_fine_direction}
\end{figure}

%% file: figures/fig_fine_position.tex
\begin{figure}[!t]
    \centering
    \includegraphics[width=0.94\linewidth]{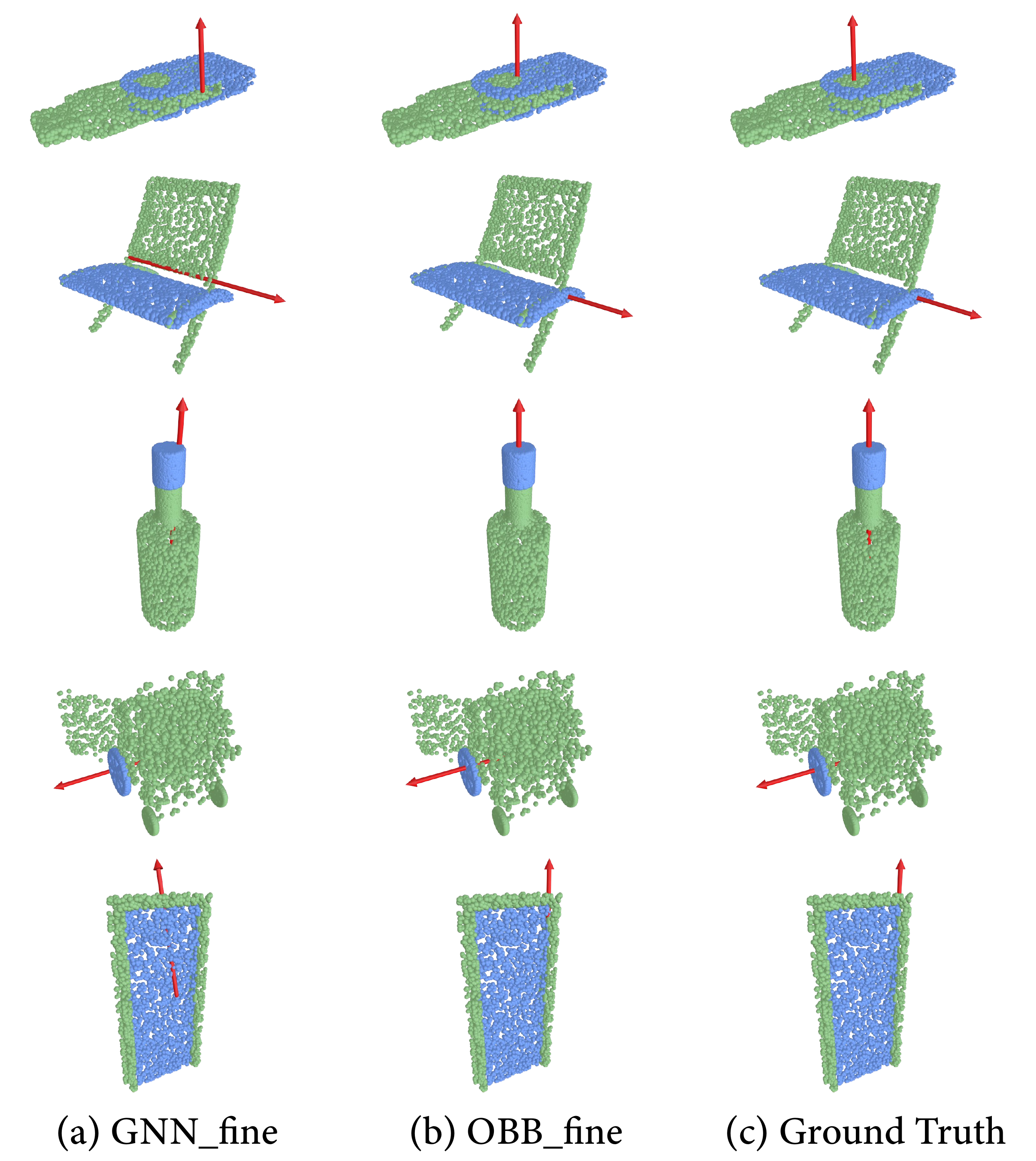}

\caption{Visual comparisons of results obtained by \emph{GNN\_fine} and \emph{OBB\_fine} on pre-segmented point clouds. 
The focused movable part is colored in blue while all other parts are colored in green.
\emph{OBB\_fine} outputs more accurate axis position thanks to the use of OBB priors and introduction of the interaction region.}
\label{fig:result_fine_position}
\end{figure}

%% file: figures/tab_semiweakly.tex
\begin{table*}[t!]
\footnotesize
\addtolength{\tabcolsep}{-2pt}
\centering
\begin{tabular}{c|c|ccc|c|cc} \toprule 
\multirow{2}{*}{Category}           & \multirow{2}{*}{Method} & \multicolumn{3}{c|}{Part-based Metrics}                                           & Joint States       & \multicolumn{2}{c}{Joint Parameters}      \\ 
                                    &                         & Rotation Error °↓         & Translation Error↓         & 3D IoU ↑                & Error↓             & Angle Error°↓      & Distance Error↓      \\ \midrule 
                              
\multirow{3}{*}{refrigerator} & ANCSH                      & 12.09 18.43 21.70         & 0.148 0.751 0.606          & 73.2 3.0 5.7            & 16.22 19.83        & 1.75 1.78          & 0.035 0.044          \\
                              & ANCSH (Semi)              & 10.41 18.39 20.87         & 0.134 0.715 0.573          & 71.5 4.8 6.6            & 18.33 21.37        & 1.70 1.72          & 0.032 0.038          \\
                              & ANCSH (Semi-weakly)        & \textbf{5.07 10.05 10.09} & \textbf{0.066 0.144 0.177} & \textbf{75.8 14.4 29.3} & \textbf{9.23 9.61} & \textbf{1.62 1.62} & \textbf{0.020 0.018} \\
 \midrule

 \multirow{3}{*}{dishwasher}   & ANCSH                      & 10.78 18.97               & 0.155 0.160                & 73.9 \textbf{40.3 }              & 16.23              & 3.21               & 0.0396               \\
                              & ANCSH (Semi)              & 12.62 21.16               & 0.178 0.208                & 70.7 38.3               & 17.84              & 3.08               & 0.0481               \\
                              & ANCSH (Semi-weakly)        & \textbf{7.94 14.88}       & \textbf{0.126 0.145}       & \textbf{74.0} 37.0      & \textbf{15.02}     & \textbf{2.67}      & \textbf{0.0349}      \\
                             
 \midrule
 
 \multirow{3}{*}{table}        & ANCSH                      & 12.27 10.27               & 0.167 \textbf{0.185}                & 67.2 \textbf{53.6}               & \textbf{0.30}      & 4.35               & -                    \\
                              & ANCSH (Semi)               & 9.17 9.15                 & 0.156 0.205                & 70.7 51.3               & 0.34               & 4.43               & -                    \\
                              & ANCSH (Semi-weakly)       & \textbf{4.13 6.43}        & \textbf{0.082}  0.237      & \textbf{70.2}  50.0      & 0.32               & \textbf{3.03}      & -                    \\
                              
 \midrule

\multirow{3}{*}{door\_set}    & ANCSH                   & 57.82 60.83                & 0.550 0.509                & 40.7 18.7               & 41.52                & 4.70               & \textbf{0.064}       \\
                              & ANCSH (semi)            & 71.05 60.97                & 0.682 0.535                & 35.7 18.3               & 54.17                & 5.43               & 0.110                \\
                              & ANCSH   (semi-weakly)     & \textbf{44.55 54.00}       & \textbf{0.4251 0.4395}     & \textbf{43.3 19.1}      & \textbf{27.17}       & \textbf{3.70}      & 0.066 \\
 \midrule
\multirow{3}{*}{scissors}     & ANCSH                      & 14.65 16.18               & 0.164 0.174                & 34.3 36.7               & 19.86              & 2.20                & 0.034                \\
                              & ANCSH (Semi)               & 14.20 16.18               & 0.162 0.172                & 33.7 36.0               & 18.87              & 2.32               & 0.029                \\
                              & ANCSH (Semi-weakly)       & \textbf{12.00 13.52}      & \textbf{0.121 0.143}       & \textbf{37.2 37.8}      & \textbf{14.80}     & \textbf{2.10}       & \textbf{0.026}     
                             
                             \\ 
\midrule
\multirow{3}{*}{mean}     & ANCSH   & 23.66  &	0.31 &	42.00 &	19.19 	& 3.25 &	0.04       \\
                          & ANCSH (Semi) & 24.76 & 0.32 & 41.00 & 22.21 & 3.39 & 0.06     \\
                         & ANCSH (Semi-weakly)       & \textbf{17.43} & \textbf{0.20} &	\textbf{44.85} &	\textbf{13.35} &	\textbf{2.62} &	\textbf{0.04} \\ \bottomrule

\end{tabular}
\caption{Quantitative evaluation of ANCSH used in different settings for motion prediction on partial point clouds, including \emph{ANCSH} for the original fully-supervised setting, \emph{ANCSH (Semi)} for the semi-supervised setting, and  \emph{ANCSH (Semi-weakly)} for our semi-weakly supervised setting.
The results show that \emph{ANCSH (Semi-weakly)} yields the best performance.}
\label{tab:result_semiweakly}
\end{table*}

%% file: figures/fig_result_partial.tex
\begin{figure*}
    \centering
    \includegraphics[width=0.94\linewidth]{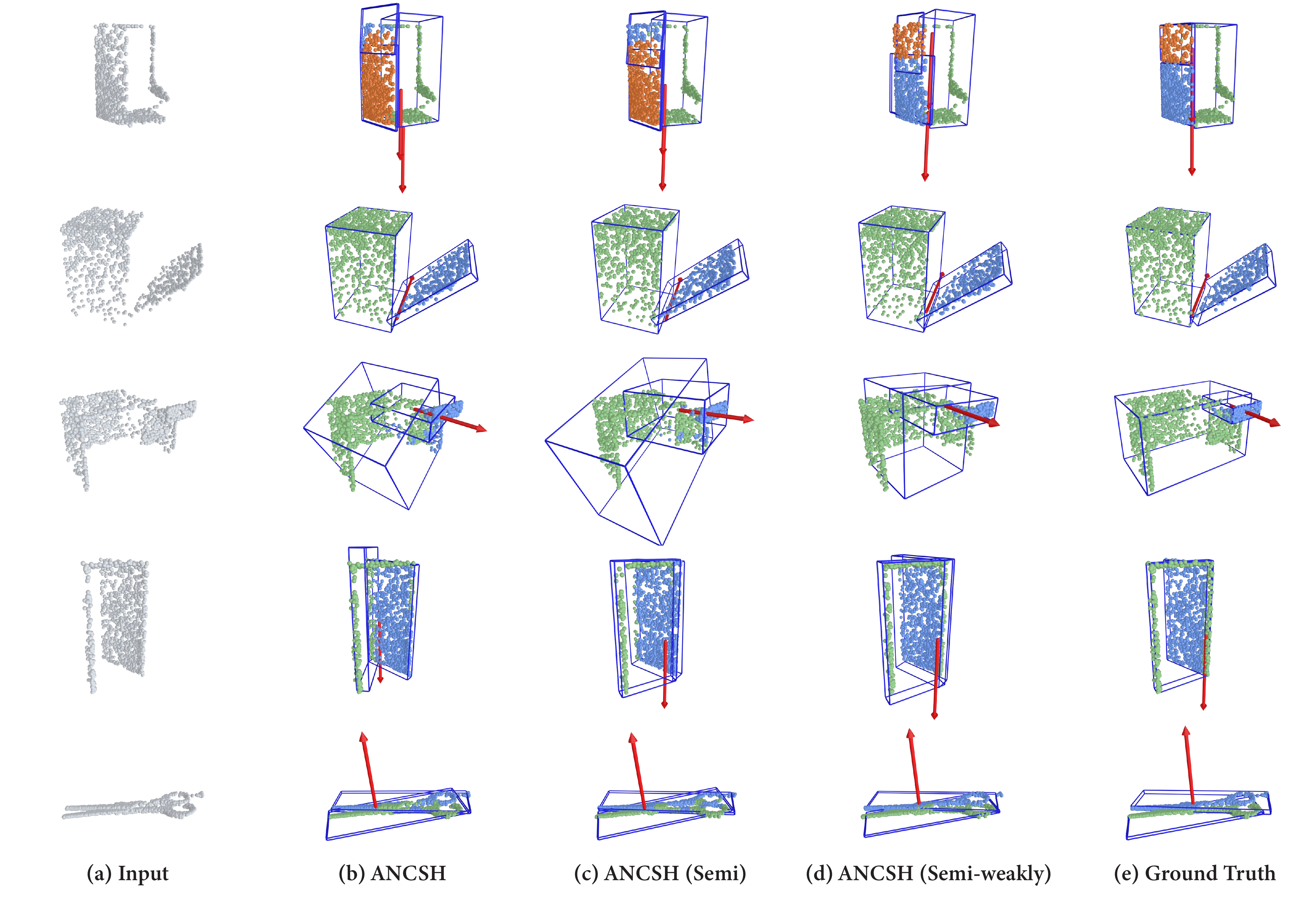}

\caption{Visual comparisons of results obtained by ANCSH \cite{li2020category} when trained in different settings for motion prediction on partial point clouds, with \emph{ANCSH} for the original fully-supervised setting, \emph{ANCSH (Semi)} for the semi-supervised setting, and  \emph{ANCSH (Semi-weakly)} for our semi-weakly supervised setting.
Note how the results have been improved when trained in our semi-weakly supervised setting. }
\label{fig:result_partial}
\end{figure*}

%% file: figures/tab_supervision.tex
\begin{table*}
\footnotesize\addtolength{\tabcolsep}{-2pt}
\centering
\begin{tabular}{c|c|ccc|c|cc} \toprule 
\multirow{2}{*}{Category}           & \multirow{2}{*}{Method} & \multicolumn{3}{c|}{Part-based Metrics}                                           & Joint States       & \multicolumn{2}{c}{Joint Parameters}      \\ 
                                    &                         & Rotation Error °↓         & Translation Error↓         & 3D IoU ↑                & Error↓             & Angle Error°↓      & Distance Error↓      \\ \midrule

\multirow{5}{*}{refrigerator} & Semi-weakly (100\%)           & \textbf{5.07   10.05 10.09} & \textbf{0.066 0.144 0.177} & \textbf{75.8 14.4 29.3} & \textbf{9.23 9.61} & \textbf{1.62 1.62} & \textbf{0.020 0.018} \\
                              & Semi-weakly (80\%)            & \textbf{7.45 14.10 12.61}   & \textbf{0.096 0.171 0.188}          & \textbf{75.0 14.6 29.6}          & \textbf{12.62 10.98}        & 1.81 1.81          & \textbf{0.024 0.0192}         \\
                              & Semi-weakly (50\%)            & \textbf{12.86 23.37 18.63}    & \textbf{0.147 0.464 0.430}         &     70.1 6.6 17.8       & \textbf{20.23 17.90}          &     2.99 2.88               & \textbf{0.033 0.031}            \\
                              & Semi-weakly (30\%)            & 16.11 24.36 27.71           & 0.177 \textbf{0.516} \textbf{0.476}          & 71.7 \textbf{6.8 17.9}          & 18.59 22.11        & 1.83 1.82          & 0.039 \textbf{0.034}          \\ \cmidrule{2-8}
                              & Supervised (100\%)            & 12.09 18.43 21.70         & 0.148 0.751 0.606          & 73.2 3.0 5.7            & 16.22 19.83        & 1.75 1.78          & 0.035 0.044          \\
 \midrule

\multirow{5}{*}{scissors}     & Semi-weakly (100\%)           & \textbf{12.00 13.52}        & \textbf{0.121 0.143}       & \textbf{37.2 37.8}      & \textbf{14.80}     & \textbf{2.10}      & \textbf{0.026}       \\
                              & Semi-weakly   (80\%)          & 14.71 16.26                 & \textbf{0.142 0.165}       & \textbf{36.8 37.9}               & \textbf{18.38}              & \textbf{2.11}               & \textbf{0.027}                \\
                              & Semi-weakly (50\%)            & \textbf{13.53} 18.17         & \textbf{0.162} 0.194                & 33.5 32.2               & 20.53              & 2.28               & 0.037                \\
                              & Semi-weakly (30\%)            & \textbf{14.58} 18.29                 & 0.173 0.193                & 34.8 \textbf{35.5}               & \textbf{18.38}              & 2.38    & \textbf{0.033}         \\ \cmidrule{2-8}
                              & Supervised (100\%)                      & 14.65 16.18               & 0.164 0.174                & 34.3 36.7               & 19.86              & 2.20                & 0.034                \\ 
 \midrule
\end{tabular}
\caption{Impact of different ratios of training data for supervision. We compare the performance \emph{ANCSH} trained using 100\% labeled data and \emph{ANCSH (Semi-weakly)} trained using  different ratios of labeled data, denoted as \emph{Supervised (100\%)} and \emph{Semi-weakly (*)}, respectively.
Numbers for the semi-weakly supervised setting that outperform the 100\% supervised setting are bold.
Note that our \emph{ANCSH (Semi-weakly)} is  comparable to \emph{Supervised (100\%)} even when only half of the labeled data is used.}
\label{tab:result_supervision}
\end{table*}

%% file: figures/tab_dsg1.tex
\begin{table}[t]
\footnotesize
\addtolength{\tabcolsep}{-4pt}
\centering
\begin{tabular}{c|ccc|c|cc} \toprule 
 Method & Rota. Err.   & Trans. Err.   & 3D IoU     & Joint Err.  & Ang. Err   & Dis. Err.      \\ \midrule 
    ANCSH   & \thead{40.0 \\ 47.4 43.7}    & \thead{ 0.40 \\ 0.40 0.37}  	 &	\thead{ 62.5 \\ 20.7 17.4}   & 	\thead{30.9 \\ 27.1} 	& \thead{9.2 \\ 6.3} &	\thead{0.063 \\ 0.061}  \\
    \thead{ANCSH\\(DSG) }& \thead{ \textbf{40.0  } \\ \textbf{45.5 42.0} } &    \thead{ \textbf{0.34 } \\ \textbf{0.37 0.30} }  &  \thead{ \textbf{66.6} \\ \textbf{23.0 21.4} }  & \textbf{\thead{30.8 \\ 25.7}} & \textbf{\thead{1.5 \\ 1.5}}  &  \textbf{\thead{0.061 \\ 0.056}}   \\ \bottomrule 

\end{tabular}
\vspace{-1mm}
\caption{
Quantitative results of ANCSH for motion prediction on partial point clouds with and without using shapes from DSG-Net~\cite{yang2022dsg_rebuttal} for data augmentation.
}
\label{tab:result_dsg1}
\vspace{-6mm}
\end{table}

%% file: conclusion.tex
\section{Conclusion}
\label{sec:concl}
In this paper, we make a first try of semi-weakly supervised learning for 3D object kinematic motion prediction. We successfully transfer the labeled mobility information of limited examples from the PartNet-Mobility dataset to the PartNet dataset by leveraging fine-grained hierarchical 3D part information.
With the associated part information, our method can generate high quality pseudo motion label and also take advantage of PartNet with large-scale 3D shapes. While our model may produce incorrect predictions when the spatial relationship between the moving part and the reference part is ambiguous. For example, when the door is closed, even with correctly predicted segmentation and motion type, our method may still predict the wrong rotation axis as there are various axis distributions in our dataset. We show some representative failure cases in Figure \ref{fig:failure_cases}, where the predicted axis is shown in red while the GT axis is shown in green for comparison in each example. 

We further apply the augmented data for motion prediction of 3D partial scans and achieve a significant performance improvement. We believe such a semi-weakly supervised learning framework can be used for other motion related task and worth exploring.
\input{figures/fig_failure_cases}

%% file: figures/fig_failure_cases.tex
\begin{figure}
    \centering
    \includegraphics[width=\linewidth]{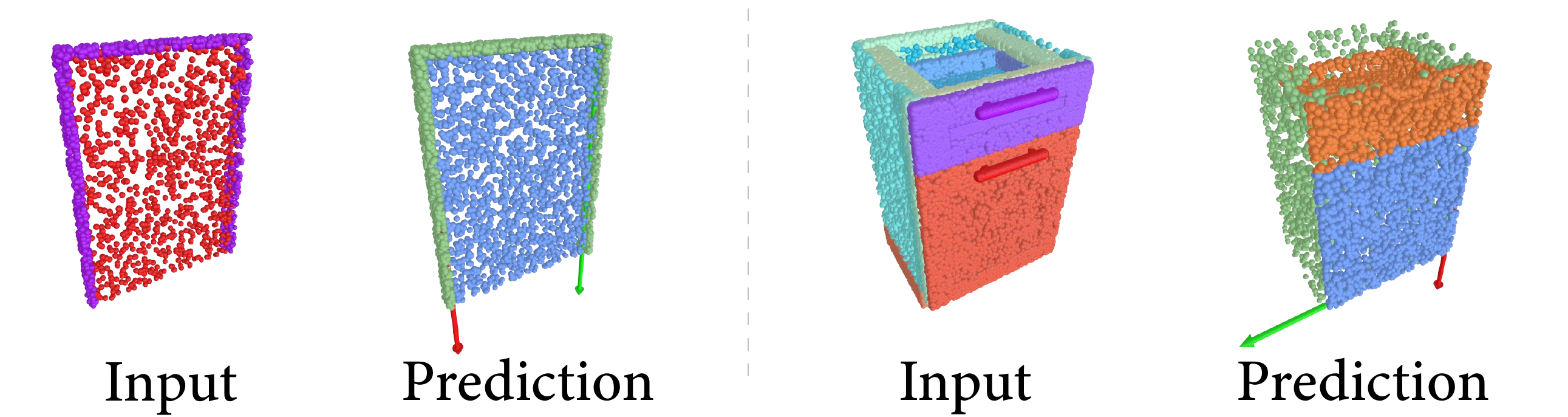}

\caption{
Failure cases of motion prediction with  \emph{ OBB\_fine}.
}
\label{fig:failure_cases}
\end{figure}